\documentclass[letterpaper]{article}
\usepackage{aaai}
\usepackage{times}
\usepackage{helvet}
\usepackage{graphicx}
\usepackage{hyperref}
\usepackage{courier}
\usepackage{natbib}
\usepackage{subfiles}
\usepackage{multirow} % Required for multirows
\usepackage{longtable} % To display tables on several pages
\usepackage{siunitx} % Required for alignment
\usepackage{booktabs} % For prettier tables
\usepackage{hhline}
\usepackage{array}
\begin{document}
%\bibliographystyle{IEEEtran}
% proceedings, working notes, and technical reports.https://www.overleaf.com/project/61660d3dd7316e181767f897https://www.overleaf.com/project/61660d3dd7316e181767f897
%
\title{Toward Defining a Domain Complexity Measure Across Domains}
% or: Toward a Transdisciplinary Domain Complexity Measure
% or: Toward a Transdiciplinary Domain Complexity Level Estimation
\author{
         % Authors
         Katarina Doctor\textsuperscript{\rm 1}, 
         Christine Task\textsuperscript{\rm 2}, 
         Eric Kildebeck\textsuperscript{\rm 3}, 
         Mayank Kejriwal\textsuperscript{\rm 4}, 
         Lawrence Holder\textsuperscript{\rm 5}, 
         Russell Leong\textsuperscript{\rm 2} 
%}
%\affiliations {
\\
\\
         % Affiliations
         \textsuperscript{\rm 1} Navy Center for Applied Research in Artificial Intelligence, Naval Research Laboratory, Washington, D.C. \\
         \textsuperscript{\rm 2} Knexus Research Corporation, Springfield, VA 
         \textsuperscript{\rm 3} University of Texas at Dallas, TX \\
         \textsuperscript{\rm 4} USC Information Sciences Institute, CA 
         \textsuperscript{\rm 5} Washington State University, WA
\\
katarina.doctor@nrl.navy.mil, \{christine.task, russell.leong\}@knexusresearch.com, \\
eric.kildebeck@utdallas.edu, kejriwal@isi.edu, holder@wsu.edu 
}
\maketitle
\begin{abstract}
\begin{quote}
Artificial Intelligence (AI) systems planned for deployment in real-world applications frequently are researched and developed in closed simulation environments where all variables are controlled and known to the simulator or labeled benchmark datasets are used. Transition from these simulators, testbeds, and benchmark datasets to more open-world domains poses significant challenges to AI systems, including significant increases in the complexity of the domain and the inclusion of real-world novelties; the open-world  environment contains numerous out-of-distribution elements that are not part in the AI systems' training set. Here, we propose a path to a general, domain-independent measure of domain complexity level. We distinguish two aspects of domain complexity: intrinsic and extrinsic. The intrinsic domain complexity is the complexity that exists by itself without any action or interaction from an AI agent performing a task on that domain. This is an agent-independent aspect of the domain complexity. The extrinsic domain complexity is agent- and task-dependent. Intrinsic and extrinsic elements combined capture the overall complexity of the domain. We frame the components that define and impact domain complexity levels in a domain-independent light. 

Domain-independent measures of complexity could enable quantitative predictions of the difficulty posed to AI systems when transitioning from one testbed or environment to another, when facing out-of-distribution data in open-world tasks, and when navigating the rapidly expanding solution and search spaces encountered in open-world domains.

\end{quote}
\end{abstract}

%=====================================================
% Keywords: open-world , real-world, out-of-distribution, domain complexity, complexity measure, complexity estimation, novelty 

%=====================================================
\section{1. Introduction}

\noindent % Introduction comes here. The purpose of this work. 
When designing AI systems that can operate in open-world settings, it is important to be aware of the complexity level of the domain for which the AI system is built and the complexity level of the domain where it will be applied or transitioned.

Self-driving cars, which have been developed and evaluated on closed courses or in simpler highway-driving contexts, will encounter new challenges safely navigating open-world surface streets \cite{Claussmann2017}. Machine learning (ML) tools in medical contexts may encounter unexpected edge cases, noisy or distorted sensor data, and potentially significant differences in data distributions when transitioning from academic benchmark data to real-world use \cite{Holzinger2017}. 

In general, agents in simulated environments navigate a much smaller set of possible states and perform deliberative-reasoning search tasks over a much smaller set of possible state-action paths than what happens in the open world of a non-simulated, real environment. As one example of an AI agent transitioning from a closed-world domain to an open-world domain, Wilson et al. (\citeyear{Wilson2014}) developed a motion and task-planning system for autonomous underwater vehicles (AUV) using a common simulation platform that they moved from development to deployment. If one takes the AUV agent, which has been built within a less complex simulated environment, installs it on a robot, and lowers it off the side of a ship into the real world, it will have significant problems correctly navigating in that real-world state space. 

In the ever-changing open-world domains, there are a plethora of novelties that will have effects on an AI agent that's been trained in a closed-world setting. The overall complexity level also changes when transitioning from a closed world to an open world. Transitioning from a closed world to an open world does not necessarily mean transitioning from a simple system to a complicated or complex system. However, understanding the boundaries of the domains we are facing will help with accommodating novelties in an open world. Knowing the complexity levels of both the domain from which an AI agent is transitioning and the domain to which that agent is transitioning will help predict the difficulty of detecting, characterizing, and accommodating novelty, and therefore will enable robustness in that agent when performing a task in the domain of interest.

The core concept of complexity levels transcends the boundaries of a domain. In this paper, we define a framework for estimating the level of complexity of domains in a transdisciplinary, domain-independent light.

%=====================================================
\section{2. Motivation and Background} 
%[Definitions and explanations of some terms that might be used differently in different fields and context.]

%\subsection{Motivation}

The Science of Artificial Intelligence and Learning for Open-world Novelty (SAIL-ON) Defense Advanced Research Projects Agency (DARPA) program \cite{Senator2019} 
solicitation states that SAIL-ON will research and develop the underlying scientific principles, general engineering techniques, and algorithms needed to create AI systems that act appropriately and effectively in novel situations that occur in open-world domains, which is a key characteristic needed for potential military applications of AI. The focus is on novelty that arises from violations of implicit or explicit assumptions in an agent’s model of the external world, including other agents, the environment, and their interactions. Specifically, the program will: (1) develop scientific principles to quantify and characterize novelty in open-world domains, (2) create AI systems that act appropriately and effectively in open-world domains, and (3) demonstrate and evaluate these systems in multiple domains.
The SAIL-ON program is divided into two groups: (a) those that facilitate the evaluations by providing novelty generators in the chosen domains across levels of the novelty hierarchy (see Table \ref{fig:Novelty Hierarchy table}), and (b) those that develop agents that can detect, characterize, and accommodate novelty. In each phase of the program, it is expected that the domain-independent characterization of novelty will be improved and that increasingly sophisticated and effective techniques for recognizing, characterizing, and responding to novel situations across domains will be developed. The novelties are categorized in a hierarchy representing the fundamental elements that make up an open-world domain, and by definition, every element within an open-world domain must have characteristic attributes and must be represented in some way. The open-world novelty hierarchy levels are: object, agent, actions, relations, interactions, rules, goals, and events (Table \ref{fig:Novelty Hierarchy table}). SAIL-ON performers have refined the novelty levels by adding difficulty levels: easy, medium, and hard. These difficulty levels refer to novelty detection and accommodation. We are treating these hierarchy levels as part of the components to consider when estimating the complexity levels of domains, which we describe in Section 5 of this paper.

\begin{table}[ht]
    \scriptsize
    \centering
    \begin{tabular}{c|c|c|p{0.6\linewidth}}
        \toprule
        \multicolumn{4}{c}{\textbf{open-world  Novelty Hierarchy}}\\
        \midrule
        {} & {Phase 1} & 1 & {\textbf{Objects:} New classes, attributes, or representations of non-volitional entities.}\\ \hhline{|~|-|-|-|}
        \multirow{-1}{.15\linewidth}{Single Entities} & {} & 2 & {\textbf{Agents:} New classes, attributes, or representations of volitional entities.}\\ \hhline{|~|~|-|-|}
        {} & {} & 3 & {\textbf{Actions:} New classes, attributes, or representations of external agent behavior.}\\ \hhline{|-|~|-|-|}
        {} & \multirow{-2}{*}{Phase 2} & 4 & {\textbf{Relations:} New classes, attributes, or representations of static properties of the relationships between multiple entities.}\\ \hhline{|~|~|-|-|}
        \multirow{-1}{.15\linewidth}{Multiple Entities} & {} & 5 & {\textbf{Interactions:} New classes, attributes, or representations of dynamic properties of behaviors impacting multiple entities.}\\ \hhline{|-|-|-|-|}
        {} & {} & 6 & {\textbf{Rules:} New classes, attributes, or representations of global constraints that impact all entities.}\\ \hhline{|~|~|-|-|}
        \multirow{-1}{.15\linewidth}{Complex Phenomena} & \multirow{-1}{*}{Phase 3} & 7 & {\textbf{Goals:} New classes, attributes, or representations of external agent objectives.}\\ \hhline{|~|~|-|-|}
        {} & {} & 8 & {\textbf{Events:} New classes, attributes, or representations of series of state changes that are not the direct result of volitional action by an external agent or the SAIL-ON agent.}\\
        \bottomrule
    \end{tabular}
    \caption{Open-world  novelty hierarchy levels developed by SAIL-ON Novelty Working Group. See acknowledgments for full list of contributors. }  \label{fig:Novelty Hierarchy table}
\end{table}

Complexity levels of domains are estimated in different ways for different domains, and it is challenging to generalize complexity estimation across multiple diverse domains. The structure of the SAIL-ON program highlights this need, for which we would like to be able to compare both domains and their complexity levels as well as to estimate the complexity level of generated novelties, which will enable of predicting the difficulty of detecting, characterizing, and accommodating novelty.

\paragraph{Theoretical Frameworks for Open-World Learning}
Some theoretical frameworks have been proposed for open-world learning. \citeauthor{Langley_AAAI_2020} (2020) proposes a framework for characterizing open-world environments with goal-directed physical agents and how those environments can change over time. \citeauthor{Boult_AAAI_2021} (2021) propose a framework for defining theories of novelty across domains. This framework measures novelty based on dissimilarity measures in the world space and the observation space. These frameworks describe novelty in open-world learning in an agent's performance perspective, but do not offer an assessment about the complexity of a domain of the novelty, which is the purpose of our paper. In this paper, we distinguish between agent-independent and -dependent parts of the domain and define the domain complexity components from both perspectives.

\paragraph{From Complexity to Difficulty} 
Understanding the complexity level of a domain helps assess and predict the difficulty of detecting, characterizing, and accommodating a novelty in that domain. Complexity and difficulty are different mental operations. The complexity level of a domain will affect the difficulty of detecting novelty, characterizing it, and accommodating it. It also will influence the difficulty of generating novelty.
``Complexity" is the description of a state --- the space of possibilities --- whereas ``difficulty" relates to the challenge posed by a specific novelty within this space of possibilities. Estimating and understanding the complexity level of a domain is an important prerequisite for a robust transition between domains and to an open-world learning.

%=====================================================
\section{3. Benefits of Knowing Domain Complexity}
 
In the Wilson et al. (2014) example, the simulated environment assumed that there was a 1-to-1 mapping between perception and the state space; i.e., that a change in perception was due to a meaningful change in state. Instead, in real-world contexts, with noise and complex factors related to environment, there may be many different perception values that all might indicate effectively the same state for the goal reasoner. The real-world perception state space is effectively much larger than the space over which goal reasoning actually is performed. An AI system that was developed and tested on a simple, simulated perception space will tend to fail when one puts it into the open sea, incessantly reporting discrepancies between its sensors and its model of the world, potentially getting trapped in a loop of constant replanning. As a result, the robot can freeze, drift off course, and lose contact with its humans. This can be a very expensive problem.  

Addressing this issue required the development of new logic for dealing with the true complexity level of the perception space. Perception information was processed with bounding boxes to reduce its complexity sufficiently so that the existing goal-reasoning and planning logic could operate over it effectively. This measure allowed the robot to operate successfully in the open water. It is important to note that this issue was task-independent. There was effectively no nontrivial task that the robot could execute successfully before its algorithm had been modified to address the increase in domain complexity compared with the simulator domain. 

Outside of the context of real-world robotics with deliberative reasoning \cite{Ingrand2013}, the domain complexity problem arises in other application areas as well. In game AI, the performance of Monte-Carlo Tree Search (MCTS) will depend on the size and complexity of the game tree; if the space of possible states and actions becomes excessively large, then the probabilistic exploration of the tree will have a higher probability of failing to sample the optimal paths, and then the agent may select moves that are suboptimal, poor, or even absurd. If the complexity of the game is understood correctly during AI development, then modifications can be made to improve the algorithm performance, such as using domain knowledge to bias search in games with large branching factors \cite{Chaslot2008}.

More dangerously, the problem arises in self-driving vehicles. \citeauthor{Claussmann2017} (2017) exhaustively surveys, categorizes, and evaluates diverse approaches to autonomous driving, but limited to only highway environments, and with respect to only eight simple maneuvers (such as changing lanes or exiting). This analysis does not cover important complexities in the highway domain, such as encountering an obstacle in the road or another vehicle merging into your lane without seeing you. Furthermore, it obviously will not apply to city street environments, which have far more states and complex transitions. Techniques that perform well in the simple, limited, highway domain may have very different properties and limitations than techniques that excel in a more open, real-world environment, and failing to understand the impact of domain complexity on algorithm choice could lead to suboptimal decision-making and potentially fatal consequences.

The issue of correctly addressing real-world domain complexity also arises in data science contexts. In medical applications, explainable AI systems increasingly are deployed to assist test analysis and decision-making \cite{Holzinger2017}. Systems that were designed for simple data distributions over one or two educational-benchmark datasets will fail or become extremely sensitive to data preparation and parameter tuning; as the feature set grows more complex, evaluation becomes more rigorous, or the data distribution becomes more diverse and heterogeneous. Again, systems developed on simple, toy-research domains, when moved to much larger, open-world domains without consideration for the change in complexity of the problem definition, will have failures with real-world consequences \cite{Holzinger2017}.  

%=====================================================
\section{4. Perspectives on Complexity from Different Disciplines}
We briefly review existing approaches to understanding and measuring complexity as it applies to relevant computational disciplines: classical AI, data science, and systems research. These existing perspectives are important to take into consideration, but none provides the comprehensive, domain-independent complexity level necessary to support the development and transition of AI to open-world domains.

\paragraph{Classical AI}
 \citeauthor{Hernandezorallo2010} (2010) use the term ``environment complexity" to refer to increasingly complex classes of domain/tasks pairs. Their thesis is that more intelligent agents are able to succeed in more complex environments. Their complexity level increases dependent on the domain's number of possible states, the number of transitions (agent actions that change the state), and the difficulty of reaching the objective (in general, winning the game). The more possible states and the more difficult-to-select actions to navigate optimally through them, the more complex the problem. 

 Relating domain complexity to the state transition graph is a common approach in classical AI systems.  \citeauthor{Ingrand2013} (2013), \citeauthor{Chaslot2008} (2008), and \citeauthor{Claussmann2017} (2017) give examples of agents whose performance is, in fact, dependent on the environment complexity as used in Hernandez-Orallo. These systems are sufficiently intelligent to solve problems of a certain complexity, but may fail as that complexity increases. 
 \citeauthor{Pereyda2020} (2020) proposed a theory for measuring complexity by taking a resource-requirements approach, focusing on the three spaces: task, solution, and policy. The authors are relating complexity to the minimum description length of agents necessary to achieve different levels of performance on the task in a domain.
 
 While these approaches are relevant to domain complexity in our context, they focus on agent-dependent, task-specific complexity and do not explicitly address intrinsic domain properties that are independent of agent and may impact solution performance independently of the task.  A domain-independent complexity level needs to take into account a wider set of components in order to fully support development for open-world domains. 
 
\paragraph{Data Science}
 Data science focuses on fitting conceptual models to input data, allowing users to make predictions about new data points. Unlike classical AI, there is no agent that can take actions to interact with its environment; instead, the domain consists of a feature set and a distribution of data points across the feature space defined by that feature set. Complexity is determined by the difficulty of fitting meaningful, accurate models to these distributions so that they support correct predictions on new data. Properties such as the size of the feature space, the sparsity of the data, and the shape of the distribution impact the difficulty of this problem. 
\citeauthor{Remus2014} (2014) consider the context of computational linguistics. They consider four task-independent domain complexity measures, focusing on the sparsity (word rarity), the feature set size (word richness), and the distribution of the input data (entropy and homogeneity). They find a strong correlation between domain complexity and the performance of a standard ML classifier on the data. Similarly to Hernández-Orallo's observations on classical AI, as complexity increases, performance decreases. These general ideas are broadly applicable as sources of complexity in problem solving, as we will discuss in Section 6. However, by themselves, they do not constitute a general definition for domain complexity levels.

\paragraph{Systems Research}
 Systems research considers complexity in the form of challenges that arise when organizing multiple interacting components, whether those components are team members, organizations, industrial production systems, software modules, or even elements of programming languages as interpreted by a compiler. In these contexts, patterns of dependencies between components are a key factor in the complexity of the problem. Large webs of interdependencies require significantly more computational time, or human cognitive load, to consider fully and to develop optimal solutions. If the interdependencies can be arbitrarily complex, then in general, the problem may be computationally hard (e.g., the knapsack problem). However, most real-world problem instances are tractable. 
 
 Various tools have been developed to help visualize, measure, and address complexity in system-design contexts \cite{Stuikys2009} \cite{Jung2020} \cite{Prnjat2001} \cite{Kim2014}. 
 
 This work relates to our problem of domain complexity, specifically in the case of domains that contain multicomponent systems. For example, when an agent must navigate a multiagent environment with a large number of external entities, it can cause a combinatorial explosion in the set of possible states and transitions necessary to capture the interactions of those entities, resulting in a dramatically larger domain. This can be an important facet of domain complexity; however, there are many other sources of complexity that do not originate from interactions with multicomponent systems. Additional components are necessary to capture the complexity of scenarios, such as single agents navigating inclement environments, ambiguous perception data, or challenging game objectives. System complexity metrics, by themselves, do not constitute a domain-independent complexity level.

%=====================================================
% Somewhere here or before motivation statememt on what is this paper about

\paragraph{ } 
In this paper, we frame the components that define and impact domain complexity levels in a domain-independent light. We organize these components into groups. We then describe methods for representing these components in a measurable form for estimating domain complexity levels.

%=====================================================
\section{5. Framework for the Components that Define a Domain's Complexity Level}
This framework is for defining the components to consider when estimating a domain's complexity level. There are two parts when determining the complexity of a domain: intrinsic and extrinsic. The intrinsic complexity is the complexity that exists by itself without any action or interaction from an AI agent performing a task on that domain; it is an agent-independent aspect of the domain complexity. The intrinsic components that define and impact the domain's complexity level are grouped further into ``environment space" and ``task solution space." The extrinsic complexity of a domain is agent-dependent and contains the ``performance space," the ``goal space," the ``planning space," and the ``skills space."

Considering the complexity level only from an intrinsic, agent-independent perspective, or only from an agent-dependent perspective, in which we only observe the complexity level of what an agent is doing and their prior knowledge, would result in a skewed metric of the complexity level. Therefore, we need to look at both the intrinsic and extrinsic parts of the domain complexity in order to get a balanced metric.  

\begin{figure}[htb]
  \centering
  \includegraphics[width=5cm]{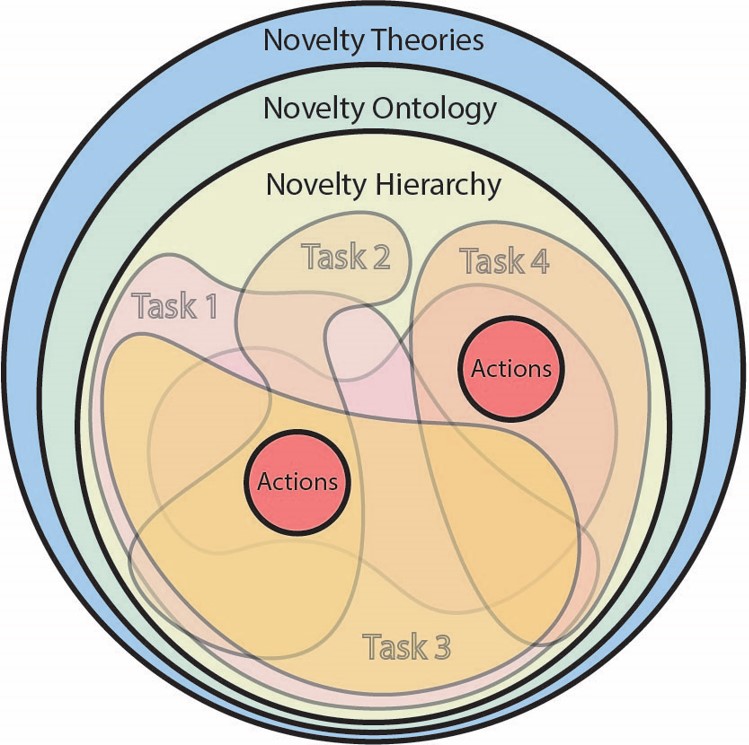}
  \caption{ Venn diagram for open-world novelty }  \label{fig:Venn-OpenWorldNovelty}
\end{figure}

We took into consideration the open-world novelty hierarchy levels described earlier in the ``Motivation" section as part of the components to consider when estimating complexity level. These hierarchy levels are: objects, agents, actions, relations, interactions, rules, goals, and events (Table \ref{fig:Novelty Hierarchy table}). We differentiate between novelty theories, ontologies, and categories: theories can describe anything that is conceptually possible, ontologies classify elements that are realistic, and categories represent a subset of elements that are practically important and scientifically useful for open-world domains. Different tasks in different domains may or may not be affected by all the hierarchy levels in the same way, as shown in Figure \ref{fig:Venn-OpenWorldNovelty}.

We list these components in groups that define the intrinsic and extrinsic domain complexity levels. Each of these components has task-dependent and task-independent parts that we do not list separately here. 

Note that these components will not be present for every domain. Also, some tasks will not interact with some of these components even if they exist in the domain.

\subsection{5.1 Components that Define Intrinsic Domain Complexity}
The intrinsic, agent-independent complexity is structural and can have many relevant levels of description for computer simulations and for the natural world. There are task-independent and task-dependent parts of the intrinsic domain complexity. The task-independent part is the input, and it is defining the structure of the domain complexity. The task-dependent part is based on which aspects of the available complexity in the domain are relevant to, or utilized in, a given task. The elements contributing to intrinsic domain complexity can be divided into the ``environment space," which includes all elements of the task environment, and the ``task solution space," which includes only those elements relevant to completing a given task within that environment.

\subsection{Environment Space}
The environment may comprise parameters/variables, data schema, tokens/observations, scale size, objects, states, or agents internal to the system. The categories of the novelty hierarchy (Table 1) represent the elements of an open-world environment space. The environment comprises discrete entities (objects and agents), and these entities can have static relationships to each other (relations) and dynamic interactions (interactions) that, in turn, can combine to create complex phenomena based on multiple relations and interactions, such as events. The complexity of the environment increases as the number of elements in each category, and the number of distinct attributes and representations for each element, increase.

\textbf{Single entities:}
The fundamental elements for an environment are the single, discrete entities present in the space, such as \textbf{objects}, \textbf{agents}, and \textbf{actions}. In action domains, this would be the number of unique objects and agents in a task environment, (e.g., blocks, tools, and enemy units in Minecraft). In perception domains, this would be the number of classes (supervised learning) or the number of clusters (unsupervised learning). Each of these categories can be expanded to include more discrete classes, and instances of each class can have an increasing number of attributes and representations. In the AUV example, an underwater simulator may have only one type of fish, with a single color and a single, swim-forward action that must be avoided by a submersible. A more complex simulator may have hundreds of types of marine life, each performing multiple actions and having variable visual appearances. Note that these agents are internal to the domain's system and not to the AI agent, who is performing a task on the domain; examples of agents that are internal in the domain are other players of a game and other drones in a swarm. Sensors that are internal in the domain would be, for example, a radar sensor that detects the AI agent's presence or movement. 

\textbf{Multiple entities:}
When multiple entities are present, new elements of the domain emerge, including static \textbf{relationships} and dynamic \textbf{interactions}. In the AUV example, in a less complex simulator, all fish may be the same size (relationship) and may swim in a straight line (interaction). In a more complex simulator, marine life may include very small and very large animals, and fish may swim in complex schooling and swarming patterns. These elements are, by definition, only present and observable when multiple entities are present in the environment. Similarly to the single-entity categories, the relationship and interaction categories can have a growing number of discrete classes, attributes, and representations as domain complexity increases. For instance, new interactions can be added, such as animals' following each other, eating each other, and fighting with each other, etc. The attributes of existing relationships and interactions also can increase in complexity, such as the speed and intricacy of schooling behavior.

\textbf{Complex phenomena:}
When multiple relations and interactions are present, complex phenomena can emerge within environments. Three examples of such complex phenomena are events, goals, and rules. 
\textbf{Events:}  When multiple interactions occur in series based on entity relationships, events can take place (e.g., a fire spreading across trees based on their physical proximity relationships). 
\textbf{Goals:}  the goals of agents in the environment also become discernible through the sequence of interactions the agent pursues within that environment. 
\textbf{Rules:} In real-world domains, and extensively in games, rules can be applied globally to define which relationships and interactions are permissible in the environment. Rules and constraints in the environment space reduce the size of the environment. These include the rules of a game, the number of states constrained by symmetry, the number of possible agent interactions, and the set of state transitions that are constrained by rules.

The following are some examples of rules: 
The rules of driving restrict which lanes can be used when driving in certain directions, the rules of Monopoly constrain spending by the amount of money the player has (0 money = 0 spending), and the rules of tic-tac-toe prevent a player from placing an X on a spot that already has an O. In AI simulations, real-world interactions or complex phenomena (e.g., a submersible requiring fuel to generate power) may be simulated as domain rules (e.g., an agent cannot issue move commands if its fuel level = 0).

For each category of elements that emerges with single entities, multiple entities, and complex phenomena, the number of distinct classes, distinct attributes, and distinct representations can be increased. Collectively, the scale and diversity of these elements determine the environment space.

\subsection{Task Solution Space} 
The task solution space comprises the number and diversity of paths that can be taken to complete a task, whether in a real, open-world domain or in a simulator, where the available world states and action space are defined. This space increases in complexity as the set of possible state transitions increases and as the available paths for success become more complex (see example state transition graphs in Figures~\ref{fig:STGExample} and \ref{fig:STGExample2}). In perception domains, the task solution space also would include the set of data-classification classes. The complexity of the task solution space is not dependent on the complexity of the environment, and the complexity of the environment may increase, decrease, or not impact the task solution space. For example, consider the domains of chess and a self-driving car in the real world. The environment of chess is quite limited, with a small number of unique objects and interactions, whereas the real world of the car may have thousands of unique objects, external agents, relationships, and interactions. If the task for the self-driving car is to move forward 1 yard, then this expansive environmental complexity has a minimal impact on the task. Furthermore, if the available action space includes a command to "move forward 1 yard," then this task becomes trivial. Winning a game of chess against a challenging opponent, by comparison, would require much more computation and strategy and would have a lower success percentage than the self-driving car's task. 

The number of possible paths, the set of possible agent interactions, and the restrictions on successful paths to achieve a goal are the primary drivers of complexity in the task solution space. We can consider a maze task in a grid environment. If the number of available paths increases from 2 to 10, then the complexity of the task solution space will increase if there is only one correct path, but will not become more challenging if every available path leads to the goal. Further, the addition of a new object class, such as a boulder, may increase the complexity of the maze environment while decreasing the complexity of the task solution space by blocking off incorrect paths and reducing the search space. 

The task solution space defines the number of possible actions and the distribution of paths (number of paths, number of intersections) through the state transition graph, the number and degree of dependencies and connections between agents and state transitions, and the degree of available strategies (defined as the set of all possible decision sequences through the environment, without violating any constraints). This set of paths can be represented as a state transition graph as explored in Section 6.

\subsection{5.2 Components that Define Extrinsic Domain Complexity}
The components of extrinsic domain complexity define the AI agent that is performing a task on the domain of interest and the skills needed to perform these tasks. The agent-dependent, or extrinsic, complexity is a mental model of an agent that has structural, observational, and planning components and is a subset of full domain complexity. 

\subsection{Performance Space} %task scoring function
Performance space is the agent-policy scoring function and the performance of the agent acting on the domain. These are the scoring of games, such as win/loss, win/loss/draw, or the range of score numbers. This also would be the reward function in reinforcement learning. 
For an AUV, the performance space would be, for example, the time taken to get from point A to point B.

\subsection{Goal Space} 
This is the number of elements or the size of the possible goals to have in the domain. Some domains have only one goal, e.g., to win, to destroy, or to move from point A to point B. Other domains have more possible goals, e.g., to win with the highest amount of money, to move from point A to B while having x amount of fuel left, and to avoid actions m, n (e.g., car: crashing, toll roads, flooded roads; aircraft: storm cloud or all clouds in VFR flight conditions).
The goal space includes the set of possible strategies, the number of goal states, and the set of all possible paths through the state transition graph that end in a goal state. Not all domains have a goal space.

\subsection{Planning Space}
Plan space is the set of possible plans that can be generated. Planners traditionally are given as input (1) a goal, (2) a set of
actions, and (3) an initial state. Many plans might accomplish the same goal, while a single plan might accomplish multiple goals. The size of the set of possible plans defines the planning space. Not all domains have a planning space. 

\subsection{Skills Space}
These are the components that define the skills that an AI agent would need to have to perform a task in a domain of interest. They consist of physical and mental skills. 

\textbf{Physical skills (hardware):}
These are the physical abilities of the AI agent --- the hardware. If a big rock is in the moving agent's path, then it is insignificant if the agent is a bulldozer robot, but it is significant if the agent is a vacuum-cleaner robot. Other characteristics include GPU, CPU, ROM/RAM, control problem, type, number, accuracy, and precision of sensors or activators. The robot AUV has sensors, such as a camera, a pressure sensor, and a sound sensor, etc. The camera might have three (RGB) channels and a specific resolution.

\textbf{Mental skills (software):} 
These are the software or knowledge, cognitive skills, and intuitive skills that an AI agent needs for performing the task in the domain of interest. 

This would be the ability for predicting, planning, setting up, or following goals. Other characteristics include the sizes and difficulty levels of interactions, 
actions, perception, goal states, and events.

\subsection{}
In summary, the described intrinsic and extrinsic components of the domain with the listed ``spaces" define the complexity level of the domain. 
These components and the listed ``spaces" will be each represented and estimated using the suggested methods in Section 6. The domain complexity level is expressed with a value of each ``space" described above. Some of the ``spaces" will not be applicable for some domains. Note that these components will have additional dimensions when we are considering a dynamic domain in which the complexity level becomes nonstationary. As the complexity level increases, the time scale will become more significant.

\begin{figure}[htb]
  \centering
  \includegraphics[width=8cm]{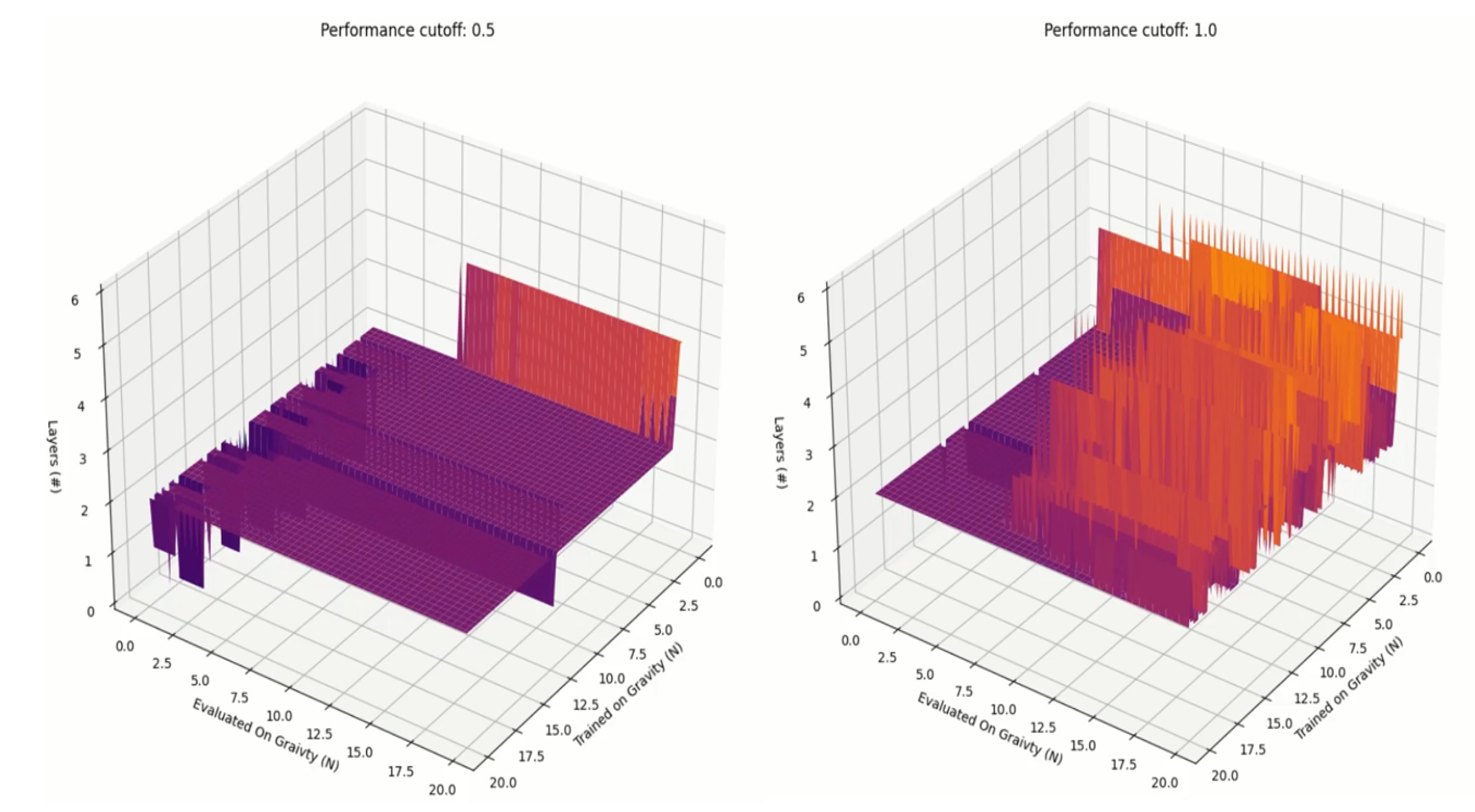}
  \caption{ Minimum number of layers needed by DQN agent to achieve (left) 50\% performance (pole balanced for 100 seconds) and (right) 100\% performance (pole balanced for 200 seconds) on tasks in which the agent is trained on one value of gravity but is tested on another value. } \label{fig:cartpole}
\end{figure}

\subsubsection{Agent-Based Extrinsic Measures}

The above dimensions define the spaces in which an agent resides in order to solve tasks in the domain. One way to identify relationships across these spaces is to define a class of agents from the skills space to solve a task and to compare the minimal complexity of an agent from the class that is capable of achieving different points in the performance space. \citeauthor{Pereyda2020} (\citeyear{Pereyda2020}) proposed a measure of task complexity that is defined as the sum of complexities of minimal-complexity agents capable of achieving the possible range of performance scores for the task. For example, Figure~\ref{fig:cartpole} depicts the complexity of a deep Q-learning agent in terms of the number of layers necessary to achieve half (0.5) and full (1.0) performance on the Cartpole task (balancing a pole on a cart by pushing the cart left and right), in which the agent is trained on one setting of gravity and is tested on another (to mimic the open-world novelty of a change in gravity). As the figure shows, more complex agents are needed to achieve higher performance, especially on tasks for which gravity increases, but less so when gravity decreases. These agent-based extrinsic results allow us to evaluate the complexity of tasks and provide insights into the intrinsic domain complexity.

%=====================================================
\section{6. Representation and Measures} % should we divide this rather into two separate chapters?
%------------------------------------------------

The perception and action/planning parts of the domains, because of their high difference, require different methods for assessing the components related to them, mentioned in the framework in Section 5. For the action and planning domains, we use state transition graphs, and for perception domains, the feature space provides a geometric representation for assessing the size of the domain and the significant components of the data distribution in the domain.

\subsection{Domain Representations}
In this section, we briefly review the two fundamental domain representations for problems in classical AI action and planning domains and data science perception domains. We then discuss several key factors of domain complexity that can be understood to hold over both of these problem contexts and domain representations. 

\begin{figure}[htb]
  \centering
  \includegraphics[width=8cm]{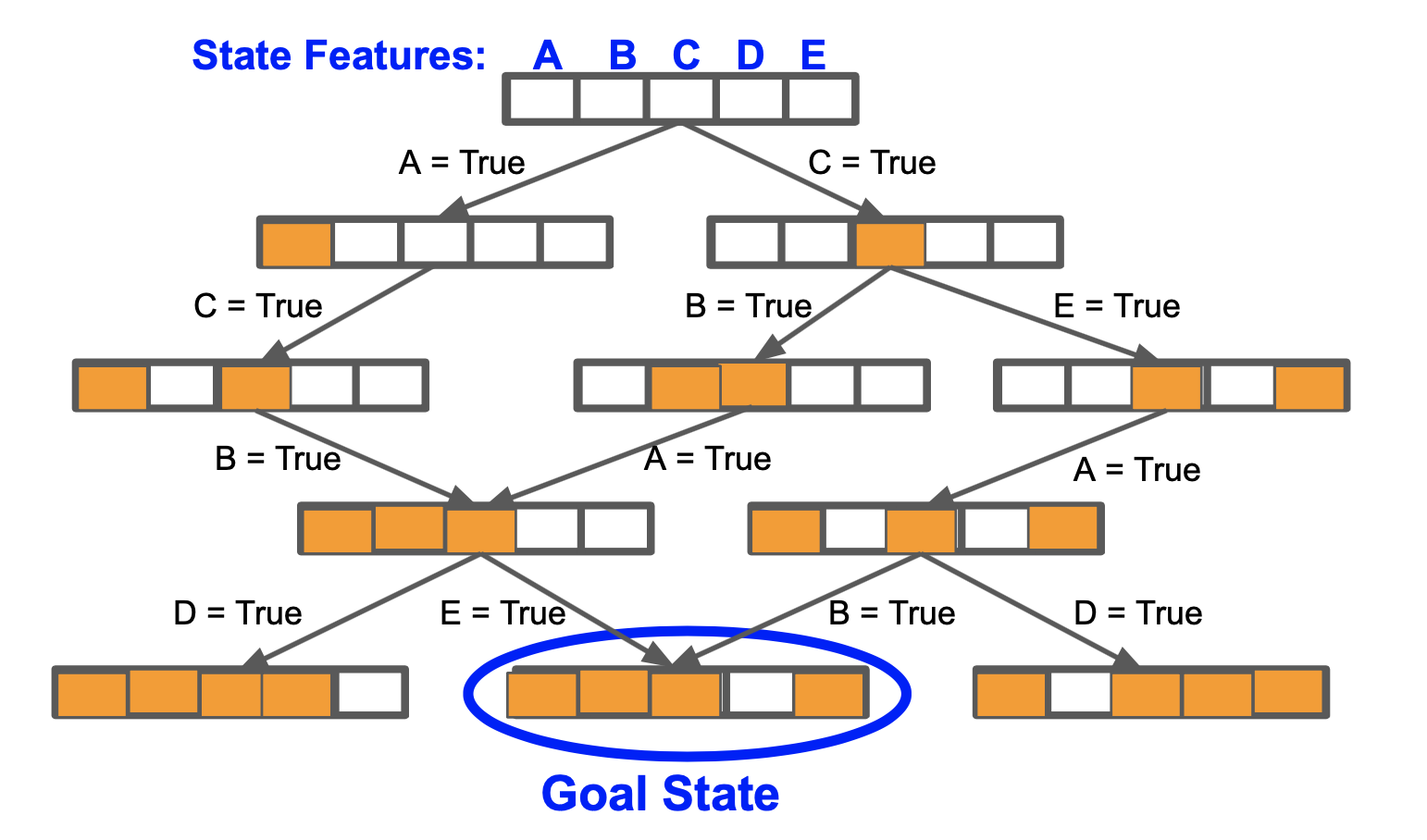}
  \caption{A simple example of a state transition graph. Possible states are depicted as nodes in the graph, connected by edges that represent actions transitioning between states. In this case, the state definition consists of five Boolean features. In the initial state, all features are set to false (white). Actions affect the state by flipping a selected feature to true (orange), and only certain actions are possible in each state. One state is highlighted as the goal state for the current task. There are relatively few possible states and up to two possible actions in each state, and multiple intersecting paths through the graph lead to the goal state. This domain and task have low complexity. } \label{fig:STGExample}
\end{figure}

\begin{figure}[!htb]
  \centering
  \includegraphics[width=8cm]{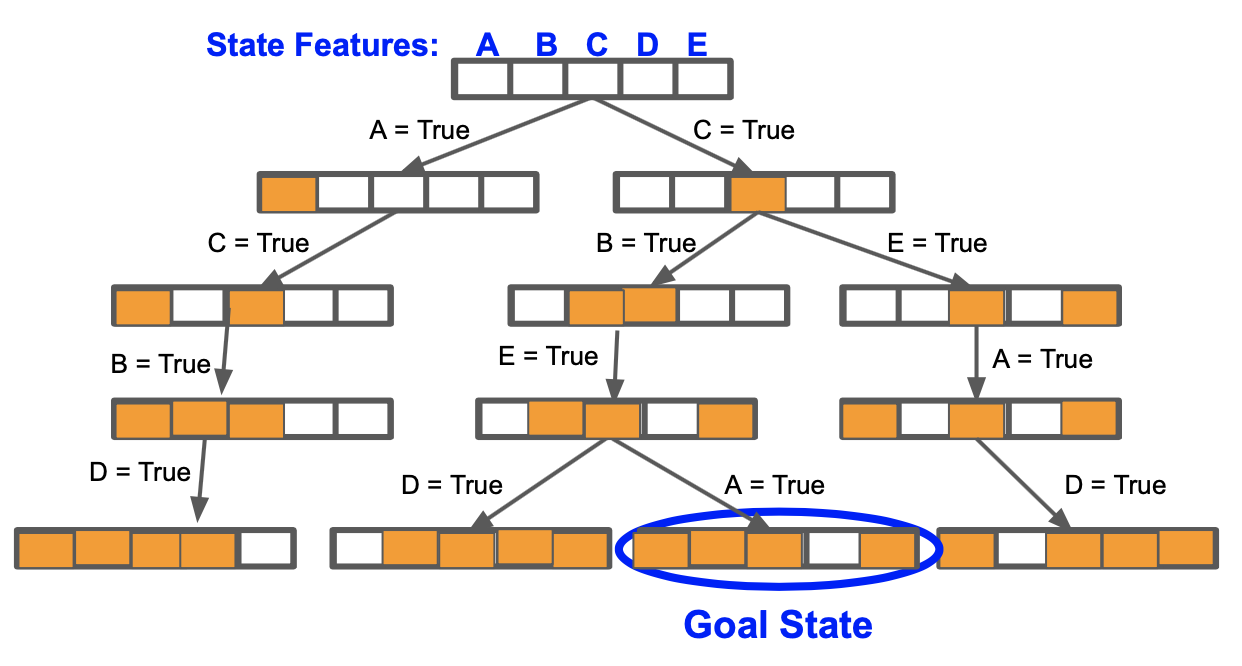}
  \caption{An example of a state transition graph for a more complex domain. Note that there is only a single path to each of the final states, meaning that if an agent makes a single incorrect decision pursuing the current task, then it will be unrecoverable and the agent will fail the task. The same holds for all states in the domain, and no matter which state is selected as the goal state, it will be challenging to achieve that goal successfully. The set of possible paths leading to a given state is more sparse in this domain than in the first domain; this domain is more complex. }  \label{fig:STGExample2}
\end{figure}

\subsubsection{State Transition Graph Space} 
A state transition graph provides a conceptual representation of the agent's domain, including possible world states, possible actions or transitions in each state, and the consequences of those actions on the state. In addition, the graph can indicate the initial state that the agent begins in at the start of execution, the set of states that satisfy the current task objective (goal states), and the possible paths that lead from the initial state to the goal state. State transitions may occur as the result of the agent's own actions, opponents' or allies' actions, or external environmental actions. Figures \ref{fig:STGExample} and \ref{fig:STGExample2} depict the state transition graph for a simple example and a more complex example.

This method of representing the domain allows us to characterize an AI action-and-planning task as identifying a potential path through the state space to the goal, successfully taking actions at each decision point to remain on an efficient, low-cost path that leads to the goal. This becomes more challenging if the complexity of the state transition graph increases; more states, more possible actions, very heterogeneous states and transitions, and sparser or more inaccessible goal paths all increase the difficulty of recognizing and maintaining an efficient, successful path to a goal state.

\subsubsection{Feature Space}
\begin{figure}[htb]
  \centering
  \includegraphics[width=5cm]{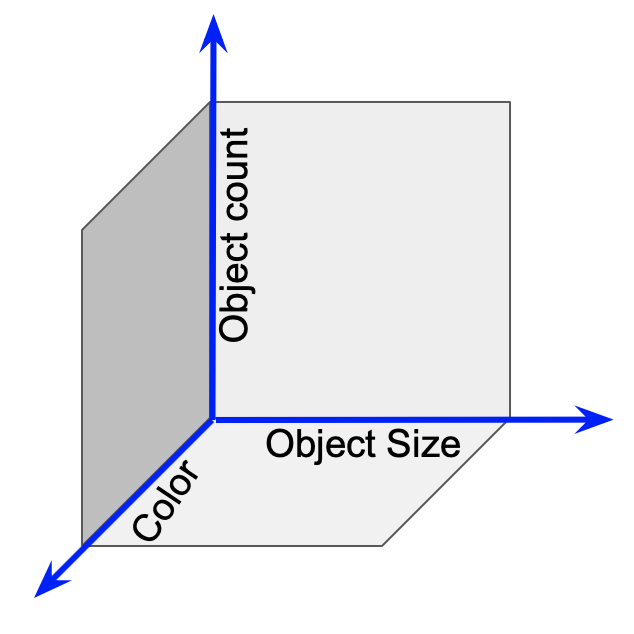}
  \caption{A 3-dimensional feature space for a dataset containing images with three features: object count, object size, and color. Each observation in the training dataset can be plotted as a point in this 3-dimensional space. ML algorithms then can characterize the distribution of points across this space (using clustering, regression, CNNs, or other techniques) so that they are able to make predictions about how new data points might fit in the existing distribution.} \label{fig:FSpaceExample}
\end{figure}

Figure \ref{fig:FSpaceExample} depicts a small, 3-dimensional feature space for a simple data science perception domain. As the number of features in a data science problem gets larger, the dimension of the feature space grows. Higher-dimensional tasks are computationally more difficult for common model-fitting techniques, such as clustering and regression. Also, as a given set of data is spread over a larger feature space, it generally grows more sparse and the distribution can become harder to characterize accurately. 

Additionally, if the data is distributed more heterogeneously across this space, then it can be difficult to characterize accurately; more diverse, outlying, or smaller subgroups of observations may not be classified correctly. Similarly to the inherent domain complexity due to the distribution of paths through the state transition graph, these are inherent difficulties with the data domain (feature space and training data distribution) that will have some impact on any classification or prediction task in that domain.  

\subsection{Complexity Measures}

\subsubsection{Dimensionality}
In both classical AI and data science applications, dimensionality impacts the difficulty of completing tasks in the domain and is a key measurement tool for estimating domain complexity. 

In classical AI, increasing the number of possible states (including increasing the environment size, the environment features, and the number of possible interactions with objects or external entities) increases the size of the state transition graph. Similarly, increasing the number of possible actions increases the breadth, or tree width, of the state transition graph. Both increase the dimensionality of the environment space and significantly increase the difficulty of tasks that involve navigating the state transition graph to reach a goal state. If a given goal state is accessible only by very long paths (which require the agent to make a long series of correct decisions), then this increases the dimensionality of the task solution and performance spaces, the complexity of the domain, and the difficulty of reaching the goal. 

In data science, increasing the number of possible features, the number of possible values for each feature (for example, by increasing the set of possible objects or environment conditions) increases the size of the feature space and the domain complexity with respect to the environment. This increases the difficulty of fitting models to characterize the distribution of the training data across that space. As vocabularies increase, sensor resolution increases, or additional variables are added, the dimensionality of the data science problem increases, and the task difficulty increases as well. Meanwhile, as the number of possible classes or prediction values increases, the set of possible parameterizations for fitting the model to the feature space can increase exponentially, increasing the complexity with regard to task solution and performance.   

%-----------------------------------

\subsubsection{Sparsity} 
In addition to dimensionality, sparsity is an important characteristic of the complexity component spaces described in the previous section. It relates to the rarity of successful strategies, or how thinly distributed the information necessary to make successful decisions can be. This is related to how difficult it is to develop successful strategies and to perform well. 

In classical AI action-and-planning domains, both the sparsity of paths that lead through the state space and the rarity of successful paths to the goal impact the complexity with regard to the task solution space and the performance space (see Figure \ref{fig:STGExample2}). When most states can be accessed through only a very small number of paths, the state space is more difficult to navigate and the domain is less forgiving of suboptimal decisions. It is easy to unintentionally enter a state from which the goal is no longer accessible. Significantly more memory or computation may be required to identify the optimal path correctly throughout the entirety of execution. 

In data science and perception domains, a sparsely distributed training dataset, especially when spread across a large feature space, significantly increases complexity with regard to the task solution space and performance. When there are many classes or prediction values that only have a few observations in the training data (i.e., many classes appear rarely in the training data) and when the training data in general is spread sparsely throughout a large dimensional feature space, there will be many modeling solutions with similar fit (given this minimal information), and it will be challenging to model the data distribution meaningfully in a way that supports accurate predictions. 

%--------------------------------------

\subsubsection{Heterogeneity} % Diversity? how complex, diverse are the concepts 
Finally, heterogeneity refers to the diversity of important information in a given domain component. Problems that include more diversity generally require more information and, therefore, more complicated solutions to address correctly. 

Classical AI action/planning problems that have a very heterogeneous state space (environment space) in which different states often have different possible action sets (and the same action may produce very different transitions, depending on the feature values of the current state) have an increased complexity with regard to the task solution and the performance space --- they require more information to navigate successfully. This often occurs in contexts in which a diverse set of skills is required to navigate the state space; domains with high skill-set complexity will have high heterogeneity. 

In data science and perception domains, heterogeneity can occur in the feature space when different features have very different properties with respect to the data distribution (i.e., combining data from different sensors in cases with high skill-set complexity). Additionally, classes or prediction values, which reflect very diverse subgroups in the data, may be more difficult to model efficiently and accurately.

%------------------------------------------
\paragraph{}Domain complexity level in a transdisciplinary light differs in different ways. Domain complexity level can indicate that the domain is simple, complicated, or complex in different ways. For this reason, we propose using a final, single value of a complexity level rather than complexity level values for the environment, policy, solution, and planning spaces described in Section 5. These four spaces' complexity level can be expressed as a bar chart or a spider chart (with its caveats). The skills space expressed in these four spaces would show what skills are needed for an agent to perform a task successfully in that domain.

%=====================================================
\section{7. Conclusion}

Returning to our example of the AUV from \citeauthor{Wilson2014} (2014), we can express the difficulty that was encountered when transitioning from the simulator domain to the real-world domain in terms of the components of our complexity levels. The real world had higher dimensionality with regard to the environment space complexity and greater heterogeneity with respect to the perception-task solution space (where noise and environmental factors meant a diverse collection of different sensor readings might represent a single system state). These unexpected increases in domain complexity, as compared to the simulator domain, caused the AI to behave pathologically, constantly replanning in response to the discrepancies between its perception of the real world and its less complex internal state space. Once the complexity of the deployment domain was understood, a bounding-box technique was introduced, enabling the robot to operate successfully in the open ocean.

%=====================================================
\section{8. Next Steps}

This paper is an early attempt at synthesizing several measures and aspects of domain complexity that are hypothesized to apply generally across domains and tasks. Many questions remain. We believe that these questions need to be formulated and addressed within a research agenda that treats domain-\emph{general} complexity (as opposed to domain-\emph{specific} complexity) as a first-class citizen. We are not claiming that each such research question should apply empirically to every domain, but ideally, it should be applicable to a diverse set of domains, allowing general claims to be made. 
As next steps, we state three questions below that (potentially) could be investigated in an experimental setting with the goal of yielding general insights about domain complexity:

\begin{enumerate}
    \item Can we derive strong theoretical connections between domain complexity and \emph{difficulty}? For instance, will it always be the case that tasks in a more complex domain \emph{necessarily} will be more difficult, or just more difficult \emph{in expectation}? And can difficulty be studied independently of complexity? These are questions with which multiple communities, especially within AI and open-world learning, are only just beginning to grapple. In addition to stating theoretical claims, it is also important to test these claims empirically. We believe that designing appropriate experimental methodologies that allow us to validate general claims is, in itself, a promising area for future research to tackle.
    \item Can our measures of domain complexity be used to define and \emph{quantify} complexity in so-called complex systems, such as networks and dynamical systems, as well as other nonlinear systems (e.g., differential equations)? Is one system more complex than another, and if so, then along what dimensions? Moreover, does this have theoretical ramifications, validated by appropriate experiments, for analyzing such systems? 
    \item In the cases of infinite action and state spaces, what are the appropriate mathematical frameworks for distinguishing between domains of (arguably) differing complexity? In real analysis, for instance, there are different hierarchies of infinity that are well understood. For instance, integers and real numbers both form infinite sets, but the latter has provably greater cardinality than the former. Could similar claims be made for complexity?
\end{enumerate}

These questions are not exhaustive, and some have several other questions associated with them that may require several parallel lines of theoretical and experimental research to investigate fully. We emphasize that the one commonality between all these questions is their lack of dependence on a single domain or model. Rather, all of them aim toward an agenda that is as domain-independent as possible. As noted earlier, we believe that this is the central element that distinguishes other field-specific studies of complexity from our proposal.

\section{Acknowledgments}
This research was  sponsored  by DARPA and the Army Research Office (ARO) under multiple contracts/agreements, including  
% Eric, UTD contract#, 
W911NF2020010, 
% Mayank, USC-ISI contract#, 
W911NF2020003, and  
% Larry, WSU contract#.
W911NF-20-2-0004.
 The views contained in this document are those of the authors and should not be interpreted as representing the official policies, either expressed or implied, of DARPA, ARO, or the U.S. government.

The following teams and individuals contributed to the open-world novelty hierarchy presented in Table 1: Washington State University led by Lawrence Holder, Australian National University led by Jochen Renz, University of Southern California led by Mayank Kejriwal, University of Maryland led by Abhniav Shrivastava, University of Texas at Dallas led by Eric Kildebeck, University of Massachusetts at Amherst led by David Jensen, Tufts University led by Matthias Scheutz, Rutgers led by Patrick Shafto, Georgia Tech led by Mark Riedl, PAR Government led by Eric Robertson, SRI International led by Giedrius Burachas, Charles River Analytics led by Bryan Loyall, Xerox PARC led by Shiwali Mohan, Smart Information Flow Technologies led by David Musliner, Raytheon BBN Technologies led by Bill Ferguson, Kitware led by Anthony Hoogs, Tom Dietterich, Marshall Brinn, and Jivko Sinapov.
%\newpage

\bigskip
\noindent

\nocite{*}
% \printbibliography
\bibliography{sources.bib}
\bibliographystyle{aaai}
%\printbibliography
\end{document}